\documentclass[journal,twoside,web]{ieeecolor}
\usepackage{generic}
\usepackage{cite}
\usepackage{amsmath,amssymb,amsfonts}
\usepackage{algorithmic}
\usepackage{graphicx}
\usepackage{multirow}
\usepackage{textcomp}
\usepackage{color}
\usepackage{array}
\usepackage{booktabs}
\usepackage[marginal]{footmisc}
\usepackage{bm}
\usepackage{bbding}
\usepackage{tabularx}

\newcommand{\tabincell}[2]{\begin{tabular}{@{}#1@{}}#2\end{tabular}}  

\begin{document}
\title{Margin Preserving Self-paced  Contrastive Learning Towards Domain Adaptation for Medical Image Segmentation}
\author{Zhizhe Liu,
	Zhenfeng Zhu$^{*}$,
	Shuai Zheng,
	Yang Liu,
	Jiayu Zhou,
	and Yao Zhao,~\IEEEmembership{Senior Member,~IEEE,}
\thanks{Z. Liu, Z. Zhu, S. Zheng, Y. Liu and Y. Zhao are with the Institute of Information Science, Beijing Jiaotong University, Beijing 100044, China, and also with the Beijing Key Laboratory of Advanced Information Science and Network Technology, Beijing Jiaotong University, Beijing 100044, China (e-mail: zhzliu@bjtu.edu.cn; zhfzhu@bjtu.edu.cn; zs1997@bjtu.edu.cn; 19112005@bjtu.edu.cn; yzhao@bjtu.edu.cn). (\emph {Corresponding author: Zhenfeng Zhu.}).}
\thanks{Jiayu Zhou is with the Department of Computer Science and Engineering, Michigan State University, East Lansing, MI, 48823. (e-mail: jiayuz@cse.msu.edu).}}

\maketitle

\begin{abstract}
To bridge the gap between the source and target domains in unsupervised domain adaptation (UDA), the most common strategy puts focus on matching the marginal distributions in the feature space through adversarial learning.
However, such category-agnostic global alignment lacks of exploiting the class-level joint distributions, causing the aligned distribution less discriminative.
To address this issue, we propose in this paper a novel margin preserving self-paced contrastive Learning (MPSCL) model for cross-modal medical image segmentation.
Unlike the conventional construction of contrastive pairs in contrastive learning, the domain-adaptive category prototypes are utilized to constitute the positive and negative sample pairs.
With the guidance of progressively refined semantic prototypes, a novel margin preserving contrastive loss is proposed to boost the discriminability of embedded representation space.
To enhance the supervision for contrastive learning, more informative pseudo-labels are generated in target domain in a self-paced way, thus benefiting the category-aware distribution alignment for UDA.
Furthermore, the domain-invariant representations are learned through joint contrastive learning between the two domains.
Extensive experiments on cross-modal cardiac segmentation tasks demonstrate that MPSCL significantly improves semantic segmentation performance, and outperforms a wide variety of state-of-the-art methods by a large margin.
The code is available https://github.com/TFboys-lzz/MPSCL.
\end{abstract}
\begin{IEEEkeywords}
unsupervised domain adaptation, image segmentation, contrastive learning, adversarial learning.
\end{IEEEkeywords}

\section{Introduction}
\IEEEPARstart{S}{emantic} segmentation refers to the task of assigning a category label to each pixel in an image.
Recently, some works based on Deep Neural Networks (DNN) have gained impressive advances in medical image segmentation task, e.g., brain lesion~\cite{li2018fully}, neuronal
structures~\cite{ronneberger2015u} and so on~\cite{arezoomand20153d,chen2020unsupervised}.
Among these works, most of them~\cite{ronneberger2015u,li2018fully} are on the basis of assumption that enough labeled data is available for target task.
However, such assumption is seriously limited in many real-world clinical scenarios.
Taking the recent outbreak of epidemic as an example, we have been facing a global health crisis, i.e., the pandemic of a novel Coronavirus Disease (COVID-19)~\cite{wang2020novel,huang2020clinical}, since December 2019.
Due to the high cost of annotation and the urgent work of doctors to combat the pandemic, there is not enough annotated data to train a well-performing DNN.
One of the most common solutions is to train a model on a label-rich domain (named as source) and then generalize it to a label-lacking domain (named as target).
However, since the significant distribution gap between the two domains (i.e., domain shift problem), the trained model usually suffers from a sharp drop in performance when applied to the target domain.

\begin{figure}[t]
	\begin{center}
		\includegraphics[width=0.35\textwidth ]{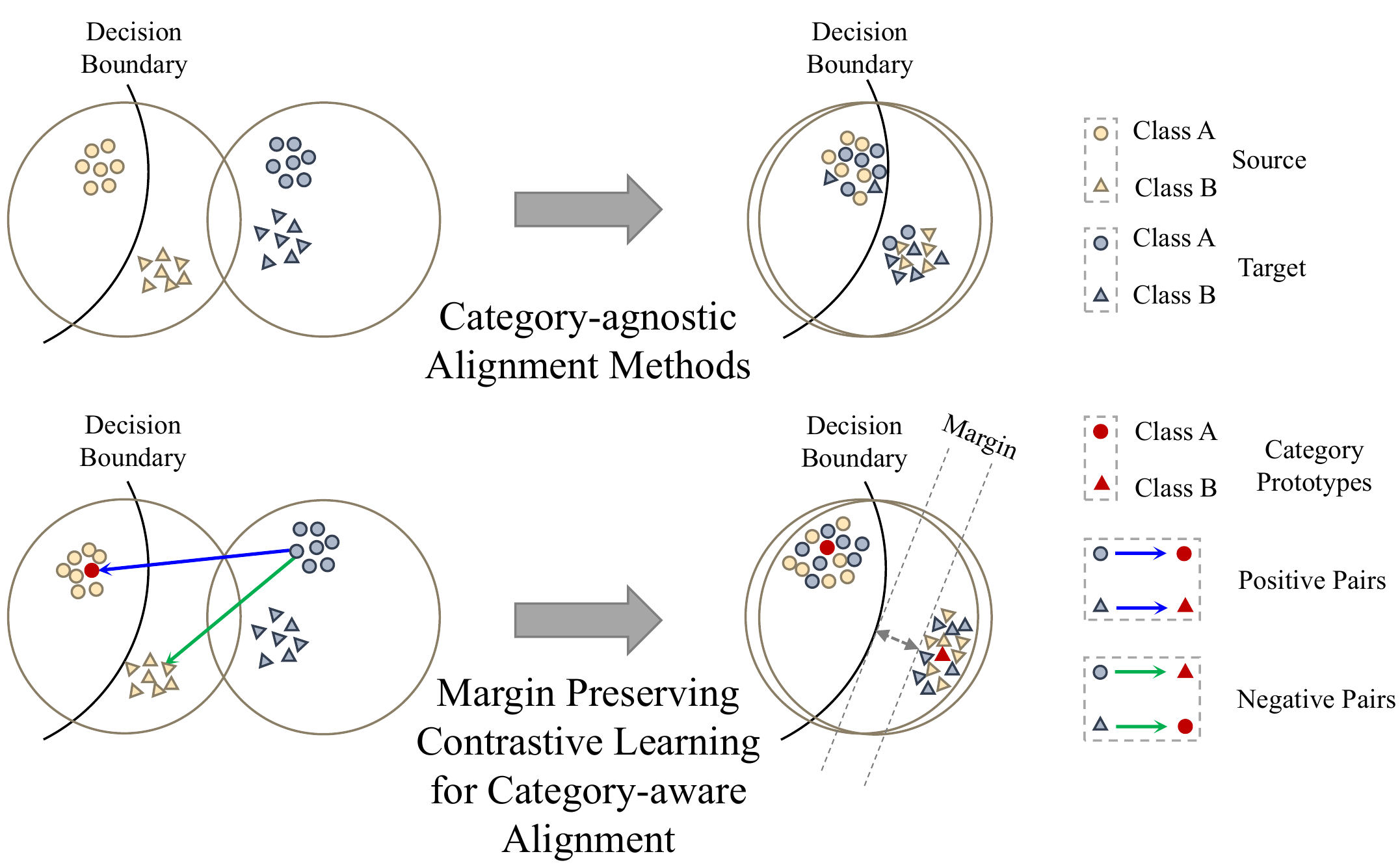}
	\end{center} 

	\caption{Category-agnostic VS. the proposed category-aware domain alignment.
		\textbf{Top:} Previous category-agnostic domain alignment methods that aim to align global marginal distributions but ignoring the semantic consistency. \textbf{Bottom:} The proposed margin preserving contrastive learning method for category-aware feature alignment. Obviously, we can boost the inter-class difference and reduce the intra-class variation by constructing enough positive and negative pairs.}
	\label{fig:intro_cl}

\end{figure}


Many approaches based on unsupervised domain adaptation (UDA) ~\cite{long2015learning,sun2016deep,tsai2018learning,vu2019advent,pan2020unsupervised,dou2018unsupervised} have recently been proposed to make the knowledge learned from the source domain better transferred to the target. 
Most previous methods employ representation learning based on some distance metrics (e.g., the maximum mean discrepancies (MMD)~\cite{long2015learning}) or adversarial learning~\cite{pan2020unsupervised,dou2018unsupervised} to bridge the gap between the two domains. 
Although the methods based on adversarial training have achieved impressive progress in domain adaptation, most of them are limited to align only the global marginal distributions of visual representations without considering semantic consistency as shown at the top of Fig.~\ref{fig:intro_cl}.
Here, semantic consistency means that after domain adaptation alignment, the distributions of the same category from different domains should be identical in the embedding space, while the distribution between categories can be easily distinguished.

\begin{figure}[!tbp]
	
	\begin{center}
		\includegraphics[width=0.32\textwidth ]{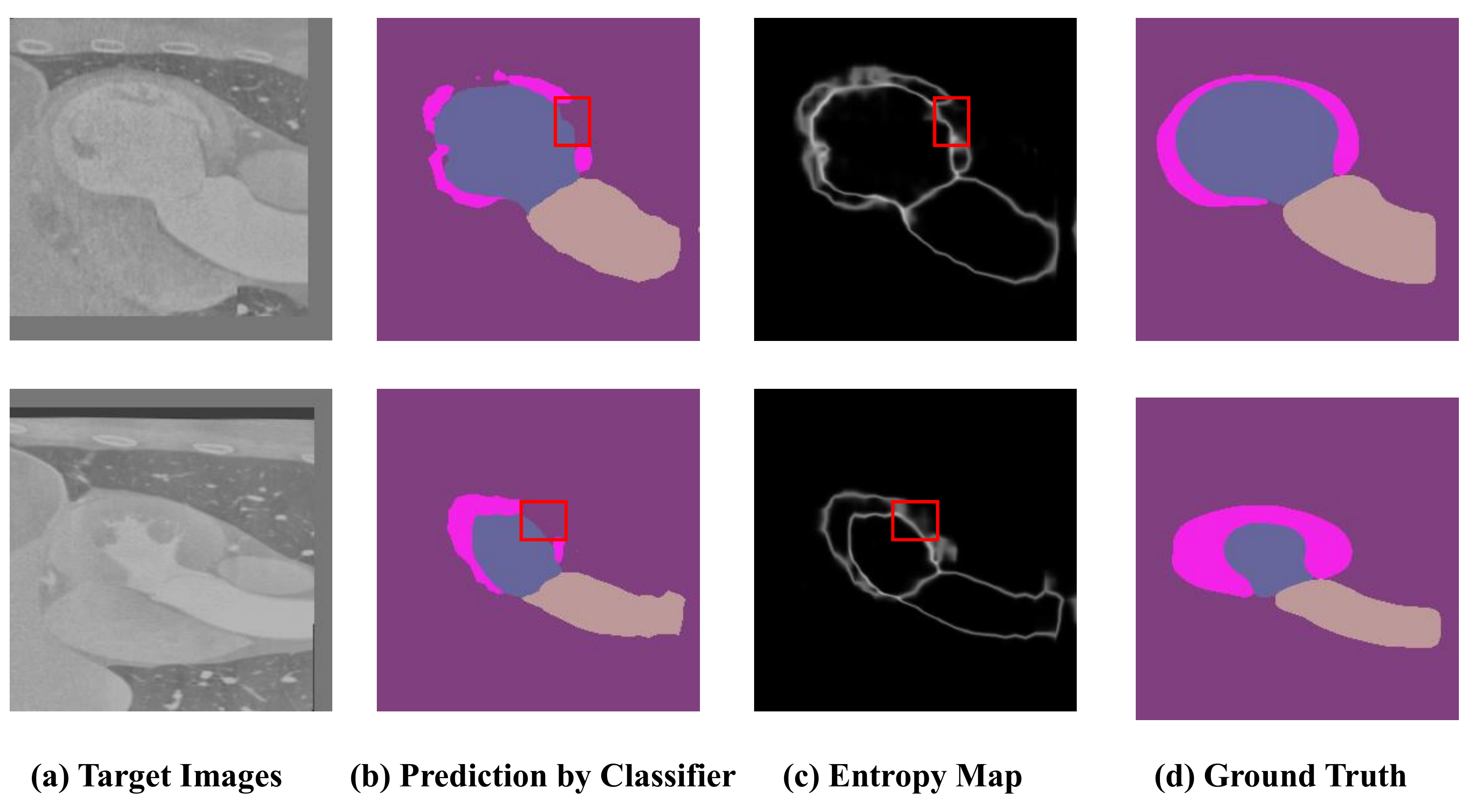}
	\end{center} 
	\caption{Illustration of the incorrect and overconfident predictions generated on the target data (i.e., marked with red bounding boxes).
		(a) The target slice images. (b) The predicted masks generated by the classifier. (c)The entropy map, given the predicted category probability vector $p\in \mathcal{R}^L$ of a pixel, the entropy value $E=\sum_{i=1}^L -p_i\log p_i$. (d) The ground-truth labels.}
	\label{fig:classify_problem}
\end{figure}

\begin{table}[!tbp]
	\scriptsize
	\renewcommand\tabcolsep{4.pt}
	\renewcommand\arraystretch{0.9}
	\caption{Comparisons of our MPSCL with the state-of-the-art UDA methords.}
	\begin{tabular}{|c|c|c|c|c|}  \hline  
		Methods & \tabincell{c} {Global domain \\alignment} & \tabincell{c} {Local category \\alignment} & \tabincell{c} {Intra-category \\compactness} & \tabincell{c} {Inter-category \\separability}\\
		\hline 
		AdaOutput~\cite{tsai2018learning} & \Checkmark & \XSolid    & \XSolid     & \XSolid \\
		AdvEnt~\cite{vu2019advent}    & \Checkmark & \XSolid    & \XSolid     & \XSolid \\
		CLAN~\cite{luo2019taking}     & \Checkmark & \Checkmark & \XSolid     & \XSolid \\
		CAG~\cite{zhang2019category}       & \Checkmark & \Checkmark & \Checkmark  & \XSolid \\
		IntraDA~\cite{pan2020unsupervised}   & \Checkmark & \XSolid    & \XSolid     & \XSolid \\
		SIFAv2~\cite{chen2020unsupervised}    & \Checkmark & \XSolid    & \XSolid     & \XSolid  \\
		MPSCL     & \Checkmark & \Checkmark & \Checkmark  & \Checkmark \\
		\hline
	\end{tabular}
	\label{tab:com_uda}
\end{table}

To avoid the semantic confusion between the two domains, some methods~\cite{zou2018unsupervised,li2019bidirectional} tried to generate pseudo-labels for target data by self-training, providing more powerful supervision for classifier training.
As an anchor-guided UDA model for semantic segmentation, both category-wise domain alignment and self-training were facilitated in an explicit way~\cite{zhang2019category}. Despite the category-wise domain alignment, the local semantic structure in the embedding space was not adequately considered~\cite{zhang2019category} thus ignoring the inter-category separability.
As a result, the inter-category difference, which is crucial for dense pixel-wise prediction task, won't be sufficiently boosted.
Fortunately, it has been shown in some recent works~\cite{misra2020self,chen2020simple,khosla2020supervised} that contrastive learning can help to learn powerful representations by taking a closer look at both inter-category and intra-category distributions. 
We aim to address the semantic inconsistency problem, while enhancing both intra-category compactness and inter-category separability.
Fig.~\ref{fig:intro_cl} illustrates a comparison between category-agnostic models and our proposed category-aware domain alignment.
In contrast to category-agnostic models, our model is more prone to inter-category margin preserving by constructing enough semantic prototype induced contrastive pairs for contrastive learning while reducing the intra-category variation.

When applying self-training to achieve category-aware feature alignment, we need to generate some pseudo-labels for target data to match the joint distributions between the two domains.
A straightforward way~\cite{zou2018unsupervised,li2019bidirectional} is to first generate pixel-wise predictions of the target data using a classifier trained on the source data.
Then, following the self-spaced learning scheme which has been found effective for gradually learning a robust model~\cite{zhang2021few,8533384,zhang2019leveraging}, a suitable selection strategy is explored to remove error-prone predictions and generate the final pseudo-labels.
However, since the target samples usually contain hard-adapted regions, especially around the boundary regions, it may generally generate unreliable and overconfident pixel predictions.
As shown in Fig.~\ref{fig:classify_problem}, not only incorrect predictions are generated, but also it takes a lower entropy value, i.e., overconfidence. Thus, it is clearly not a trivial task to design a suitable selection strategy to avoid choosing hard-adapted regions to serve as candidate for pseudo-labeling.
However, as an intuitive assumption, the visual representations belonging to the same category in the source and well-adapted target domains usually take higher similarity. It would means the well-adapted pixel regions from the target domain can be measured in the embedding space.

Motivated by the observations above, we aim to develop in this paper a contrastive learning framework for cross-modal medical image segmentation,
in which the semantic prototypes and pseudo-labels are fully exploited. In particular, we emphasize our \textbf{contributions} as follows:

\begin{itemize}
	
	\item[-] We propose a novel \underline{M}argin \underline{P}reserving \underline{S}elf-paced \underline{C}ontrastive \underline{L}earning (MPSCL) framework to tackle cross-modal medical image segmentation.
	To the best of our knowledge, it is the first attempt that contrastive learning is applied to UDA in medical image analysis.
	
	\item[-] Different from the traditional construction of contrastive pairs in contrastive learning, the domain-adaptive semantic prototypes, which are based on the prior knowledge of source domain, are exploited to bridge the two domains and constitute the positive and negative pairs for contrastive learning.
	
	\item[-] Induced by the progressively refined semantic prototypes, a novel margin preserving contrastive loss is proposed to boost the discriminability of visual representations in embedding space. Meanwhile, the domain-invariant representations are learned via joint contrastive learning between the two domains.
	
	\item[-] To perform contrastive learning in target domain without prior label available, more informative pseudo-labels are generated in target domain via self-paced scheme, which further benefits the category-aware feature alignment.

\end{itemize}

\begin{table*}[!t]
	\scriptsize
	\vspace{-0.7cm}
	\renewcommand\arraystretch{1.55}
	\centering
	\caption{Key notations.}
	\begin{tabular}{c|c}
		\hline
		\textbf{Notations} & \textbf{Descriptions} \\
		\hline
		$N^{src},N^{trg}$   & Number of source and target image data, respectively \\
		\hline
		${H,W}$ & Height and width of an image, respectively \\
		\hline
		$x^{src}_n, y^{src}_n$ & Source image and ground-truth label of the $n$-th source domain sample: $x^{src}\in \mathbb{R}^{H\times W\times 1}$ and $y^{src}_n\in \mathbb{R}^{H\times W\times L}$ \\
		\hline
		$x^{trg}_n$, $\hat{y}_{n}^{k*}$ & Image and the $k$-th pseudo-labels of the $n$-th target domain sample. $x^{trg}\in \mathbb{R}^{H\times W\times 1}$ and $\hat{y}_{n}^{k*} \in \mathbb{R}^{H\times W\times L}$\\
		\hline
		$y^{src}_n[l;i]$ & Ground-truth label for the $l$-th pixel of $x_n^{src}$ belonging to the $i$-th category, $l \in [1, H\times W]$ and $i\in [1,L]$. \\
		\hline
		$\hat{y}_{n}^{k*}[l;i]$ & Pseudo-label for the $l$-th pixel of $x_n^{trg}$ belonging to the $i$-th category of, $l \in [1, H\times W]$ and $i\in [1,L]$.\\
		\hline
		$f(x_n^{src})$,$f(x_n^{trg})$ & Feature map of $x_n^{src}$ and $x_n^{trg}$, respectively.  \\
		\hline
		$f_n^{src}[l]$,$f_n^{trg}[l]$ & Feature vector of the $l$-th pixel in $x_n^{src}$ and feature vector of the $l$-th pixel in $x_n^{trg}$, respectively.  \\
		\hline
		$C^{(k)}=\left \{c^{(k)}_1,c^{(k)}_2,...,c^{(k)}_L\right \}$ & Category prototypes at the $k$-th iteration and $c^{(k)}_i$ represents the $i$-th category.  \\
		\hline
	\end{tabular}
	\label{table:key_notations}
\end{table*}

\section{Related Work}
\subsection{Unsupervised Domain Adaptation}
Unsupervised domain adaptation aims to alleviate the domain shift problem between the source and target domains.
In terms of how to bridge the gap between the two domains,
the existing UDA methods can be divided into three categories.
The first group aims to address the above issue by transforming the image appearance between the two domains~\cite{chen2020unsupervised,zou2020unsupervised}.
For example, with the success of CycleGAN~\cite{zhu2017unpaired} in unpaired image-to-image transformation, Chen et al.~\cite{chen2020unsupervised} proposed to transform the labeled source MRI images to the appearance of target CT images and then utilized the synthesized target-liked images to train a segmentation model.
Different from the approaches based on image alignment, the other stream chooses to bridge the distribution gap between the two domains in the feature space~\cite{long2013transfer,long2015learning,luo2019taking,zhang2020collaborative}.
Specially, benefiting from the advances of generative adversarial networks~\cite{goodfellow2014generative}, which has been widely used in representation learning~\cite{8402089,zheng2020distribution}, some methods~\cite{luo2019taking,zhang2020collaborative} have focused on learning domain-invariant representations by a minimax game between a generator and a discriminator.
Inspired by the fact that the segmentation outputs of images from two domains should have considerable similarities, e.g., spatial layout and local context, many recent methods~\cite{tsai2018learning,vu2019advent,zeng2020entropy} tended to perform structure adaptation between the two domains in the output level.
Working along this line,~\cite{vu2019advent} proposed an entropy-based adversarial learning to penalize low-confident predictions on target domain.
The discrepancy between our MPSCL and the state-of-the-art UDA methods is presented in Tab.~\ref{tab:com_uda}.
It can be seen that: i) Different from some global domain alignment methods (e.g., AdvEnt~\cite{vu2019advent} and SIFAv2~\cite{chen2020unsupervised}), we also conduct local category alignment, which can further improve the transferability of the learned model. ii) Compared with some local domain alignment methods (e.g., CLAN~\cite{luo2019taking} and CAG~\cite{zhang2019category}), our model enhances not only the intra-category compactness but also the inter-category separability, thus making the learned representations more discriminative.

\subsection{Contrastive Learning}
Contrastive learning aims at learning an embedding representation space by maximizing similarity and dissimilarity on positive and negative data pairs, which has been extensively used in the metric learning~\cite{schroff2015facenet} and self-supervised learning (SSL)~\cite{wu2018unsupervised,misra2020self,chen2020simple}.
In the SSL setting, where the supervised information of training data is unavailable, contrastive learning focus on learning an invariant representation space by designing various pretext tasks based on data transformations(e.g., rotation cropping and color jittering)~\cite{misra2020self,chen2020simple}.
Recently, Khosla et al.~\cite{khosla2020supervised} have extended the contrastive loss for supervised training. Due to the exploration of local semantic structures, it is able to learn more powerful representations.
Although contrastive learning has achieved impressive results in representation learning, it performs yet poorly in cross-modal medical semantic segmentation, in which the significant distribution gap between the two domains keeps to be a hard nut to crack.
Concretely, since the supervision of target domain is unavailable, it fails to directly bridge the two domains in the semantic level and construct enough contrastive pairs.
Thus, different from the way for construction of paired data in SSL, the domain-adaptive prototypes are utilized in our MPSCL to serve as category anchors, guiding the construction of contrastive pairs in feature space for contrastive learning.

\subsection{Self-training}
Self-training, which typically includes a teacher-student framework, uses a good teacher model trained on the labeled data to assign pseudo-labels to the unlabeled data, and then utilizes human labels and pseudo-labels to jointly train a student model.
In deep learning, self-training has received increasing interest due to the dramatic reduction in cost of data labeling (e.g., image classification~\cite{xie2020self}, machine translation~\cite{he2019revisiting} and speech recognition~\cite{kahn2020self}).
Recently, some UDA works~\cite{zou2018unsupervised,zhang2019category,pan2020unsupervised} have attempted to generate pseudo-labels for the target data to achieve category-aware domain alignment.
For example,~\cite{zhang2019category} leveraged an anchor-based pixel-level distance loss to match the joint distributions between the two domains in the feature space by self-training. But, since it fails to make full use of local semantic structure information, the knowledge learned from source domain will not generalize well to the target domain.
Typically, referring the self-paced learning scheme~\cite{zhang2021few,8533384,zhang2019leveraging} which is usually used to learn a more robust model by introduce a regularizer term, these UDA methods also apply the ``easy-to-hard" training scheme, starting the training process with the most confident pseudo-labels.

\section{METHODOLOGY}

\begin{figure}[htbp]
	\begin{center}
		\includegraphics[width=0.38\textwidth ]{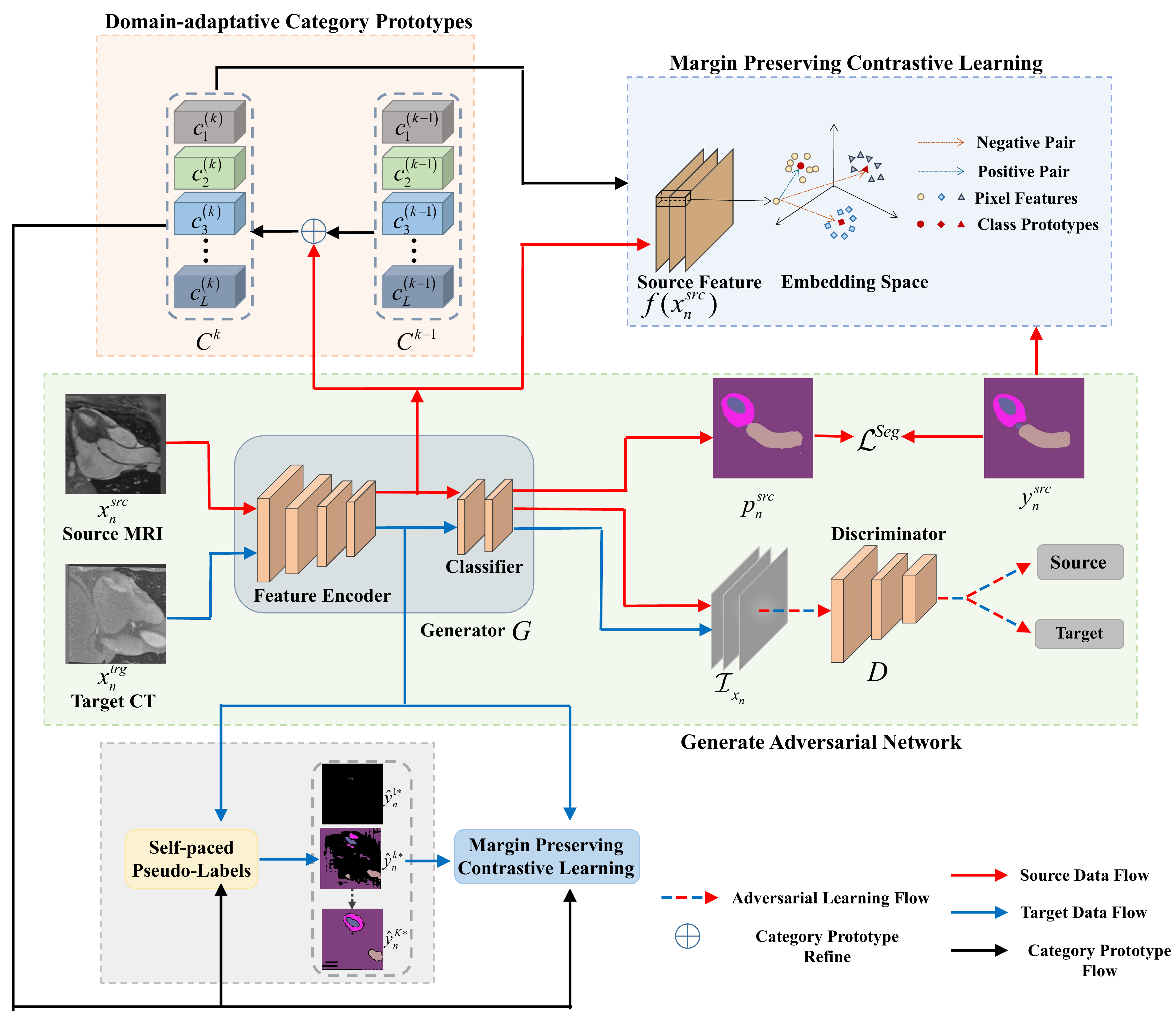}
	\end{center} 
	\caption{Overview of the proposed margin preserving self-paced contrastive learning UDA framework. $C^{(k-1)}$ and $C^{(k)}$ denote the domain-adaptive category prototypes at the $(k-1)$-th iteration and the $k$-th iteration. $p^{src}_n$ and $y^{src}_n$ represent the predicted and ground-truth masks of $x^{src}_n$. $\hat{y}^{(k)*}_n$ represents the pseudo-labels of $x^{trg}_n$ at the $k$-th iteration. $\mathcal{I}_{x_n}$ denotes the weighted self-information of $x_n^{src}$ or $x_n^{trg}$. $\mathcal{L}^{Seg}$ represents image segmentation task loss for supervised learning on the source domain.}
	\label{fig:overall_framework}
	
\end{figure}

\subsection{Problem Definition}\label{problem_defi}
The key notations used throughout this paper are summarized in Table~\ref{table:key_notations}.
Given a labeled source dataset $\left\{ X^{src},Y^{src}\right\}$, a semantic segmentation model aims to learn a mapping $\mathcal{F}$ from the image domain $X^{src}$ to the label domain $\mathcal{Y}^{src}$:
\begin{align}
\begin{split}
\mathcal{F}: X^{src} \rightarrow \mathcal{Y}^{src}
\end{split}
\end{align}
Specially, the mapping function $\mathcal{F}$ can be obtained by minimizing a hybrid loss $\mathcal{L}^{Seg}$ that is generally defined as:
\begin{equation}\label{loss:seg}
\begin{aligned}
\mathcal{L}^{Seg} = \sum_{n=1}^{N^{src}} \left(\mathcal{L}_n^{CE}(y^{src}_n,p^{src}_n) +  \mathcal{L}_n^{Dice}(y^{src}_n,p^{src}_n) \right)
\end{aligned}
\end{equation}
where $p^{src}_n\in \mathbb{R}^{H\times W \times L}$ denotes the pixel-wise prediction by $\mathcal{F}$, and $L$ is the number of categories.
In addition, the first term $\mathcal{L}_n^{CE}$ is the weighted cross-entropy loss for pixel-level classification,
and the second term $\mathcal{L}_n^{Dice}$ is the Dice loss which is usually applied in medical image segmentation tasks with multiple organ structures.
As for the design of hybrid loss $\mathcal{L}^{Seg}$, the central point is how to tackle the class imbalance in medical image segmentation~\cite{dou2018unsupervised}.
Generally, a model trained on the source domain $X^{src}$ is hard to directly generalize to the target domain due to the significant distribution discrepancies between the two domains.

Recently, several UDA methods have been proposed to bridge the gap, which can be formulated as :
\begin{align}
\begin{split}
\mathcal{F}_{uda}: X^{src} \cup X^{trg} \rightarrow \mathcal{Y}^{src} \cup \mathcal{Y}^{trg}
\end{split}
\end{align}
where $\mathcal{F}_{uda}$ is trained on the labeled source domain $\left \{ X^{src}, Y^{src} \right \}$ and unlabeled target domain $\left\{ X^{trg}\right\}$.
Typically, the mapping function $\mathcal{F}_{uda}$ aims to learn a domain-invariant representation space by distilling transferable knowledge from the source domain. Here, $\mathcal{Y}^{src}$ and $\mathcal{Y}^{trg}$ are assumed to be identical as in general setting.

\subsection{Overall Framework}\label{overall_fra}
As shown in Fig.~\ref{fig:overall_framework}, the overall framework of MPSCL model mainly contains four components, i.e., Generative
Adversarial Network, Domain-adaptive Category Prototypes, Self-paced Pseudo-labels, and Margin Preserving Contrastive Learning.
\begin{itemize}
	\item \textbf{Generative Adversarial Network} The generative adversarial network is utilized as a backbone alignment network to promote the category-aware alignment between the two domains. Particularly, the generator contains three branches, one of which generates the predicted masks of source domain for supervised learning, and the other two produce the weighted self-information maps of source and target domains for adversarial learning.
	\item \textbf{Domain-adaptive Category Prototypes} The category prototypes are exploited to constitute the contrastive pairs for the join contrastive learning between the two domains. To make these prototypes well domain-adaptive, they are refined in a progressive way in model training.
	\item \textbf{Self-paced Pseudo-labels} In order to conduct contrastive learning in target domain without prior label available, the informative self-paced pseudo-labels are generated for the target data to provide extra supervision.
	\item \textbf{Margin Preserving Contrastive Learning} To boost the discriminability of representation via generator, a novel margin preserving contrastive learning loss is proposed. While ensuring the tight clustering within categories, the difference between categories can be maximized.
\end{itemize}

\subsection{Domain-adaptive Category Prototypes}\label{prototypes}
Due to the significant distribution discrepancies between the two domains, the traditional construction methods of contrastive pairs (e.g., rotation cropping and color jittering) cannot be suitable for cross-domain contrastive learning. To this end, the domain-adaptive category prototypes are exploited to construct contrastive pairs.
Specially, to obtain the representative prototypes,  we first initialize the category prototypes $C^{(0)}=\left \{c^{(0)}_1,c^{(0)}_2,...,c^{(0)}_L\right \}$ using the category centers of the initial source pixel feature, and we have:
 \begin{align}
\begin{split}
c^{(0)}_i = \frac{1}{|{\cal N}_{i}|} \sum_{n=1}^{N^{src}}\sum_{l=1}^{H\times W}y_{n}^{src}[l;i]f^{(0)src}_{n}[l]
\end{split}
\end{align}
where $|{\cal N}_{i}|$ denotes the number of pixels belonging to the $i$-th category in source domain, i.e., $|{\cal N}_{i}| = \sum_{n=1}^{N^{src}}\sum_{l=1}^{H\times W} y^{src}_{n}[l;i]$, and $y^{src}_{n}[l;i]=1$ if the $l$-th pixel of $x^{src}_n$ belongs to the $i$-th category.

To make the prototypes receive more pseudo-supervision from the target domain in model training and so as to have better cross-domain adaptability, we update them with a progressive refinement way in each iteration. For the category prototypes $C^{(k)}$ at the $k$-th iteration, the $i$-th category prototype $c^{(k)}_{i}$ is refined by the mean vector of pixel feature belonging to the $i$-th category in the mini-batch as:
\begin{align}\label{Refinement}
\begin{split}
c^{(k)}_i  \leftarrow \alpha c^{(k-1)}_i + (1-\alpha)\cdot \frac{1}{|{\cal B}_{i}|}\sum_{n=1}^{B}\sum_{l=1}^{H\times W}y^{src}_{n}[l;i]f^{(k)src}_n[l]
\end{split}
\end{align}
where $B$ represents the batchsize, and $|{\cal B}_{i}|$ denotes the number of pixels belonging to the $i$-th category. The $\alpha \in [0,1]$ is a momentum coefficient for moving the semantic category prototypes, and $\alpha$ is empirically set as 0.2.
Meanwhile, for the source and target domains, the category prototypes are regard as category anchors and construct contrastive pairs with each pixel feature.
Then, the joint cross-domain contrastive learning is performed to learn domain-invariant representations.

\begin{figure}[!tbp]
	\begin{center}
		\includegraphics[width=0.35\textwidth ]{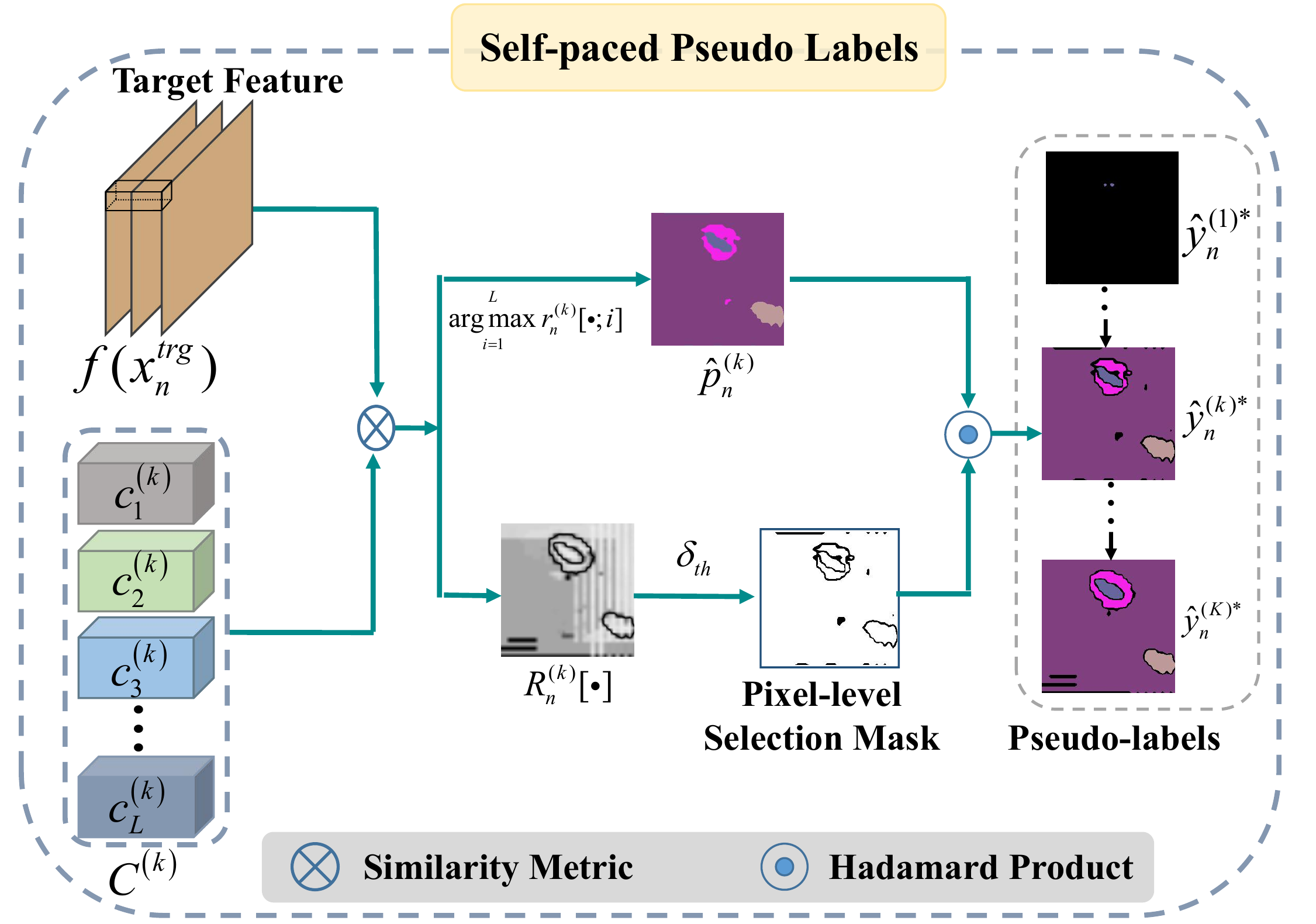}
	\end{center} 
	\caption{Illustration of the self-paced pseudo-labels. $C^{(k)}$ represents the domain-adaptive category anchor set at the $k$-th iteration. $r^{(k)}_n[\cdot;i]$ denotes the confidence scores between pixel feature and the $i$-th category anchor. $\hat{p}^{(k)}_n$ represents the predicted mask of $x^{trg}_n$ and $R^{(k)}_n[\cdot]$ denotes the confidence difference at $k$-iteration. The $\delta_{th}$ is a predefined threshold to establish a mask for interest region selection.}
	\label{fig:pseudo-labels}
\end{figure}

\subsection{Self-paced Pseudo-labels (SPL)}\label{pseudo-labels}
To conduct supervised contrastive learning in target domain without prior label available, we borrow the idea from self-training and generate pseudo-labels for the target samples. As shown in Fig.~\ref{fig:classify_problem}, it usually outputs incorrect and overconfident predictions on the target domain via the classier of generative network $G$ , especially in the early training of the model. Obviously, to assign pseudo-labels directly based on these unreliable predictions will be inevitably with high risk. To avoid selecting pixels with error-prone predictions, we propose a self-paced pseudo-labels  assigning approach in the embedding space, which is mainly based on the assumption that the well-adapted pixel feature are close to the prototype of same category and far from others.

As shown in Fig.~\ref{fig:pseudo-labels}, a self-paced selection strategy is presented by following an `easy-to-hard' scheme to capture those well-adapted pixels. Given the $l$-th pixel feature $f^{(k)trg}_n[l]$ of a target image $x^{trg}_n$ and prototype-based category anchors $C^{(k)}$, the confidence scores $\left \{r^{(k)}_{n}[l;i] = cos(\theta^{(k)}_{n}[l;i])|i=1,\cdots,L\right \}$ are first obtained by using cosine similarity:
\begin{align}
\begin{split}
cos(\theta^{(k)}_{n}[l;i]) = \frac{[c^{(k)}_i]^T f^{(k)trg}_{n}[l] }{\left \| c^{(k)}_i \right \|_2 \left \| f^{(k)trg}_{n}[l] \right \|_2}
\end{split}
\end{align}
where $\left \| \cdot \right \|_2$ denotes the $L_2$ normalization. Let's sort $\left \{r^{(k)}_{n}[l;i] |i=1,\cdots,L\right \}$ in descending order and denote $I_1$ and $I_2$ the index corresponding to the maximum and submaximum confidence scores, respectively. Hence, the final pseudo-labels are generated as follows:
\begin{equation}\label{st_solver}
\hat{y}_{n}^{(k)*}[l;\hat{y}_l]=\left\{
\begin{aligned}
1, &~\mathbf{if}~\hat{y}_l=\arg\max \limits_{i=1}^{L} r^{(k)}_{n}[l;i] \\
&\wedge ~r^{(k)}_{n}[l;I_1]-r^{(k)}_{n}[l;I_2]>\delta_{th}\\
0, &~\mathrm{otherwise}\\
\end{aligned}
\right.
\end{equation}
where $\hat{y}_l$ denotes the predicted category index of $l$-th pixel. $\delta_{th}$ is a pre-defined threshold to remove hard-adapted regions.

In practice, what is more desirable for assigning pseudo-labels is to seek those informative samples. To measure a sample whether informative or not, the confidence difference $R^{(k)}_n[l]=~r^{(k)}_{n}[l;I_1]-r^{(k)}_{n}[l;I_2]$ is adopted in Eq.(\ref{st_solver}) to characterize the significance associated to it. Correspondingly, by setting a threshold on the confidence difference, a selection mask of interest regions can be established. Since the generation pseudo-label is essentially category-prototypes induced, with the progressive refinement of category-prototype as in Eq.(\ref{Refinement}), more reliable informative pseudo-labels can be generated by the means of self-pacing, thus facilitating  the supervision for contrastive learning in target domain.

\begin{figure}[!tbp]
	
	\begin{center}
		\includegraphics[width=0.4\textwidth ]{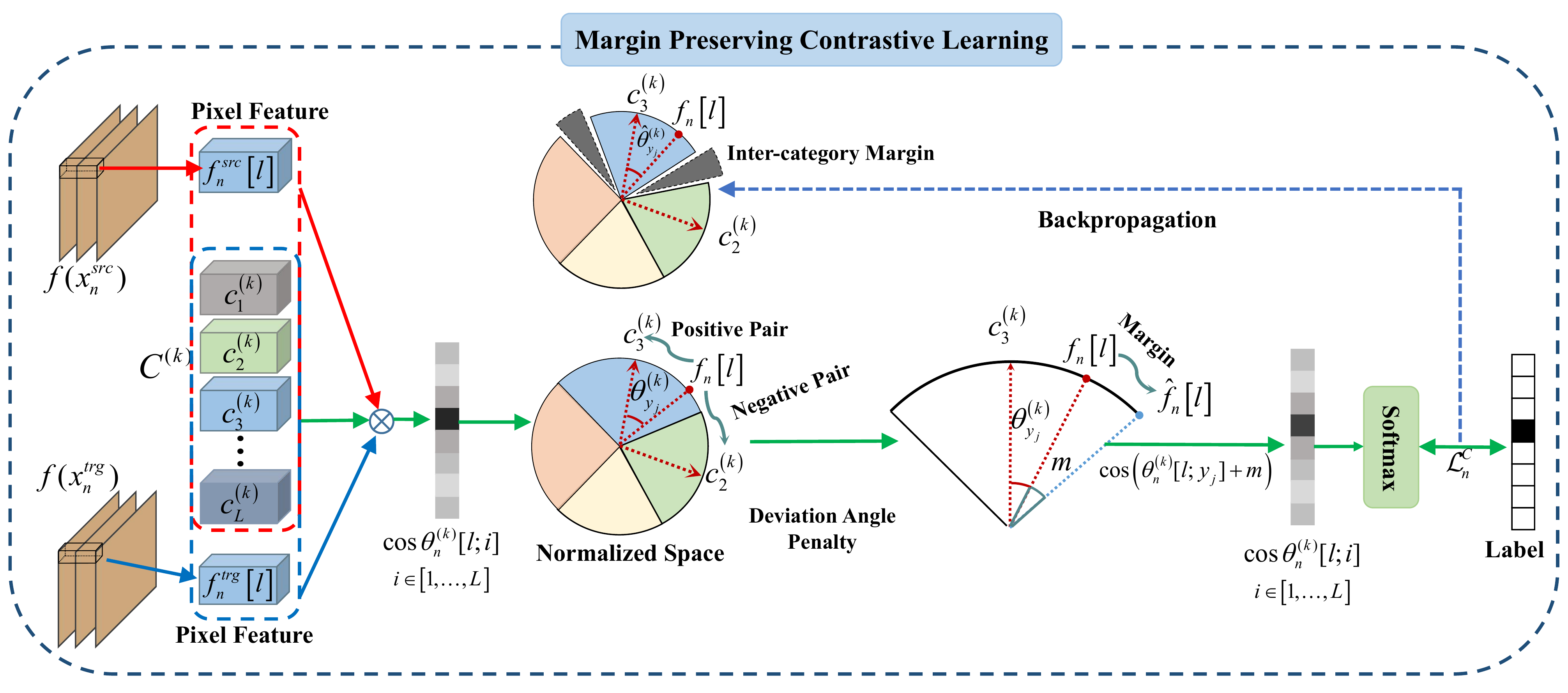}
	\end{center} 
	
	\caption{The illustration of the margin preserving contrastive loss.
		The feature of each pixel ($f^{src}_n[l]$ or $f^{trg}_n[l]$) form positive pairs with prototypes of the same categories and negative pairs with prototypes of different categories. Meanwhile, to enhance separability between categories while reducing the variation within categories, we introduce a deviation angle, i.e., $m$ in Equ. ~\ref{loss:MPCL}, as penalty to the positive pair (i.e., the pair between pixel feature and positive category prototypes). Finally, after multiple self refinement of category prototypes, the learned representation will possess distinct inter-category margin and therefore will be more discriminative.}
	
	\label{fig:MPCL}
\end{figure}

\subsection{Margin Preserving Contrastive Learning (MPCL)}\label{mpcl}
To boost the discriminability of representation, the cross-domain contrastive learning is proposed to promote the representations belonging to the same category to be closer together and far from other categories.
An intuitive way to achieve this goal is through mean square loss such as~\cite{zhang2019category}, however, it is clear that the discriminability in~\cite{zhang2019category} can not be sufficiently preserved since only the intra-category compactness is considered.
To ensure the tight clustering within categories while maximizing the difference between categories, a novel margin preserving contrastive loss is proposed.

As a geometric explanation shown in Fig.~\ref{fig:MPCL}, it can be found that there is usually a small separability between categories as well as a large variability within categories. To tackle this issue, we introduce a deviation angle as penalty to the positive anchor for margin preserving. Specifically,
the margin preserving contrastive loss of source and target domains is defined as:

\begin{equation}\label{loss:MPCL}
\mathcal{L}_n^{C} = \sum \limits_{l=1}^{H\times W}
-\log \frac{\exp( \cos (\theta^{(k)}_{n}[l;y_l] + m)/\tau )}{S[l]} \\
\end{equation}
where $\cos (\theta^{(k)}_{n}[l;y_l] )$ indicates the cosine similarity between the pixel feature and positive anchor (i.e., positive pair), $y_l$ denotes the category index of the $l$-th pixel and $S[l]=\exp( \cos (\theta^{(k)}_{n}[l;y_l] + m) / \tau) + \sum_{i=1,i \neq y_l}^L \exp( \cos (\theta^{(k)}_{n}[l;i]) /\tau)$ is a normalization and determined by the cosine similarity of each pixel.
The constant $m$ denotes the deviation angle penalty with margin preserving to positive prototype.
The temperature $\tau$ is set to avoid overfitting~\cite{hinton2015distilling}, and $\tau=1$ is a familiar setting.
As we can see from Fig.~\ref{fig:MPCL}, after some iterations via backpropagation, the inter-category difference becomes evident as the margin preserving deviation angle penalty is incorporated.
In fact, although the proposed margin preserving contrastive loss is similar to ~\cite{deng2019arcface}, we are induced by the domain-adaptive category prototypes to tackle cross-domain adaptation problem.

\subsection{Adversarial Learning}\label{adversarial}
To promote the consistency in spatial layout and local context in output space, the generative adversarial network is utilized to generate semantic masks with similar structure between the two domains.
Following~\cite{vu2019advent}, we also perform structure adaptation by minimizing the entropy via adversarial learning.
In the case of each source image $x^{src}_n$ or target image $x^{trg}_n$ as shown in Fig~\ref{fig:overall_framework}, the output of generator is applied to generate a weighted self-information map $\mathcal{I}_{x_n} \in \mathcal{R}^{H\times W \times L}$ as the discriminator input.
Here, the $\mathcal{I}_{x_n}$ is composed of pixel-level vector $\mathcal{I}_{x_n}[l;i] = - p_{n}[l;i] \log p_{n}[l;i]$, where $i=1,\cdots,L$, and $p_{n}[l;i]$ is the predicted probability of the $l$-th pixel belonging to the $i$-th category.
Similar to~\cite{goodfellow2014generative}, we let the discriminator $D$ to distinguish the input coming from the source and target domains.
Meanwhile, we train the generator $G$ to fool the discriminator $D$.
Specially, let $\mathcal{L}^B$ denote the binary cross-entropy domain classification loss, and the objective function to train the discriminator can be defined as:
\begin{equation}
\mathcal{L}^{D} = \frac{1}{N^{src}}\sum_{n=1}^{N^{src}} \mathcal{L}^B(\mathcal{I}_{x^{src}_n},1) + \frac{1}{N^{trg}}\sum_{n=1}^{N^{trg}} \mathcal{L}^B(\mathcal{I}_{x^{trg}_n},0)
\end{equation}
and the adversarial loss of the generator $G$ is:
\begin{equation}\label{loss:adv}
\mathcal{L}^{adv} = \frac{1}{N^{trg}}\sum_{n=1}^{N^{trg}} \mathcal{L}^B(\mathcal{I}_{x^{trg}_n},1)
\end{equation}
Thus, combining Eq.(\ref{loss:seg}), Eq.(\ref{loss:MPCL}) and Eq.(\ref{loss:adv}), the total optimization loss of generator $G$ is derived by:
\begin{equation}\label{loss:total}
\mathcal{L}^{G} = \mathcal{L}^{Seg} + \gamma \sum_{n=1}^{N^{src}} \mathcal{L}^{C_{src}}_n + \beta \sum_{n=1}^{N^{trg}}\mathcal{L}^{C_{trg}}_n + \lambda \mathcal{L}^{adv}
\end{equation}
where $\mathcal{L}^{C_{src}}_n$ and $\mathcal{L}^{C_{trg}}_n$ denote the margin preserving contrastive learning of source and target domains, and the $\left \{\gamma,\beta \right \} $ represent the corresponding weight factor of two domains. 
The $\lambda$ denotes the weight factor of the adversarial term $\mathcal{L}^{adv}$.

\section{EXPERIMENTAL RESULTS AND ANALYSIS}
In this section, we will present experimental results to validate the performance of our MPSCL on cross-domain semantic segmentation task.

\subsection{Dataset}
Our work in this paper mainly focus on cross-modal medical image segmentation.
Thus, the widely used Multi-Modality Whole Heart Segmentation (MMWHS) challenge 2017 dataset~\cite{zhuang2016multi} is adopted for cardiac substructure segmentation.
Specially, the training data is composed of unpaired 20 MRI and 20 CT volumes from different patient cohorts, and the ground-truth masks of these data are provided.
For evaluating our model quantitatively, we select the following four structures: ascending aorta (AA), left atrium blood cavity (LAC), left ventricle blood cavity (LVC), and myocardium of the left ventricle (MYO).
We conduct extensive experiments for cross-modal adaptation in two directions, i.e., from MRI to CT images and CT to MRI images.
For a fair comparison, we adopt the preprocessed data published by SIFAv2~\cite{chen2020unsupervised}. 


\subsection{Evaluation Metrics}
During test phase, we are mainly interested in two aspects of performance, i.e., the overlap and the difference between predictions and ground-truth masks.
Accordingly, we adopt two common metrics, the Dice similarity coefficient (Dice) and the average symmetric surface distance (ASD), to quantitatively analyze the performance of our model.
As for Dice, it measures the voxel-wise segmentation accuracy between the predicted segmentation and ground-truth labels, while ASD computes the average distances between the surface of the predicted masks and the ground-truth in 3D.
A higher Dice and a lower ASD value indicate better segmentation results.

\subsection{Implementation details}
In our experiments, DeepLabV2~\cite{chen2017deeplab} with pretrained parameters from ImageNet~\cite{deng2009imagenet} is selected as the generator $G$.
For the discriminator $D$, the PatchGAN configuration is adopted in cardiac CT to MRI task while the discriminator configuration of AdaOutput~\cite{tsai2018learning} is used in cardiac MRI to CT task.
To provide the representative category prototypes and informative pseudo-labels, we first train MPSCL (more than 4000 iterations) with $\beta = 0$, $\gamma = 0$, and $\lambda = 0.003$.
Afterwards, the domain-adaptive prototypes are initialized by the category centers of initial source feature, and the self-paced pseudo-labels are generated for the target domain.
To avoid choosing hard-adapted pixel regions and dropping useful information, the threshold $\delta_{th}$ is set as 0.25.
Then, We continually train MPSCL with $\gamma = 1.0$, $\beta = 0.1$ and $\lambda = 0.003$ and progressively refine the category prototypes and pseudo-labels.
The deviation angle penalty $m$ is selected from $\left\{ 0.2, 0.4\right\}$, and the temperature $\tau$ is set to 1.0.
Meanwhile, similarly to~\cite{tsai2018learning}, we perform domain adaptation on multi-level outputs from \textit{conv4} and \textit{conv5} to further improve performance.
During the training procedure, our model, except the discriminator, is trained using Stochastic Gradient Descent optimizer~\cite{bottou2010large} with learning rate $2.5 \times 10^{-4}$, momentum 0.9 and weight decay $10^{-4}$.
The Adam optimizer~\cite{kingma2014adam} with learning rate $10^{-4}$ is used for training the discriminator.

\begin{table}[!tbp]
	\footnotesize
	\caption{Performance comparison with different unsupervised domain adaptation methods for cardiac segmentation.
	}
	\centering
	\begin{center}
		\resizebox{0.4\textwidth}{!}{%
			\begin{tabular}{c|ccccc|ccccc}
				\toprule[1.0pt]
				
				\multicolumn{11}{c}{Cardiac MRI $\rightarrow$ Cardiac CT}\\	
				\hline
				\multirow{2}{*}{Methods} &\multicolumn{5}{c|}{Dice $\uparrow$}&\multicolumn{5}{c}{ASD$\downarrow$}\\
				\cline{2-11}
				&AA &LAC &LVC &MYO &Average &AA &LAC &LVC &MYO &Average \\
				
				\hline
				
				Supervised training &89.33 &91.36 &92.87 &88.04 &90.40 &2.27 &2.92 &1.51 &3.25 &2.49\\
				
				W/o adaptation &30.78 &36.81 &18.34 &7.21 &23.28 &20.24 &8.89 &33.57 &27.78 &22.62\\
				
				\hline
				
				AdaOutput~\cite{tsai2018learning} &73.48 &80.36 &76.14 &48.66 &69.66 &15.46 &4.81 &5.18 &6.57 &8.00 \\
				
				SIFAv1~\cite{chen2019synergistic} &81.10 &76.40 &75.70 &58.70 &73.00 &10.60 &7.40 &6.80 &7.80 &8.10\\
				
				AdvEnt~\cite{vu2019advent} &79.54&83.04 &79.54 &57.67 &74.95 &13.92 &9.30 &6.92 &4.48 &8.65\\
				
				CLAN~\cite{luo2019taking}   & 87.84 &86.79 &82.02 &60.93 &79.40 &8.23 &3.95 &4.31 &4.38 &5.22 \\
				
				CAG~\cite{zhang2019category}   & 80.70 &85.80 &80.63 &58.00 &76.28 &12.18 &4.43 &5.39 &4.28 &6.57 \\
				
				IntraDA~\cite{pan2020unsupervised}  & 48.99 &58.76 & 69.27 & 48.64 & 56.42 & 13.27 & 7.57 & 7.77 & 6.15 & 8.69 \\
				
				SIFAv2~\cite{chen2020unsupervised} &81.30 &79.50 &73.80 &61.60 &74.10 &7.90 &6.20 &5.50 &8.50 &7.00\\
				
				MPSCL  &\textbf{90.26} & \textbf{87.08} & \textbf{86.45} &\textbf{72.51} &\textbf{84.08} &\textbf{3.47} &\textbf{3.16} &\textbf{2.85} &\textbf{3.41} &\textbf{3.47}\\
				
				\bottomrule[1.0pt]
		\end{tabular}}
	\end{center}
	
	\centering
	\begin{center}
		\resizebox{0.4\textwidth}{!}{%
			\begin{tabular}{c|ccccc|ccccc}
				\toprule[1.0pt]
				
				\multicolumn{11}{c}{Cardiac CT $\rightarrow$ Cardiac MRI}\\	
				\hline
				\multirow{2}{*}{Methods} &\multicolumn{5}{c|}{Dice $\uparrow$}&\multicolumn{5}{c}{ASD$\downarrow$}\\
				\cline{2-11}
				&AA &LAC &LVC &MYO &Average &AA &LAC &LVC &MYO &Average \\
				
				\hline
				
				Supervised training &81.65 &86.33 &92.29 &80.02 &85.07 &3.43 &2.09 &1.66 &1.63 &2.20\\
				
				W/o adaptation &18.52 &7.25 &53.54 & 2.08 & 20.35 &7.07 &25.81 &8.65 &29.97 &17.87\\
				
				\hline
				
				AdaOutput~\cite{tsai2018learning} &52.32 &71.79 &79.54 &49.26 &63.23 &6.01 &3.56 &5.07 &4.38 &4.76 \\
				
				SIFAv1~\cite{chen2019synergistic} &67.0 &60.70 &75.10 &45.80 &62.10 &6.20 &9.80 &4.40 &4.40 &6.20\\
				
				AdvEnt~\cite{vu2019advent} &54.40&72.01 &77.49 &51.75 &63.91 &6.79 & 3.21 & 3.92 &4.03 &4.48\\
				
				CLAN~\cite{luo2019taking}   & 39.01 &57.62 &74.60 &50.57 &55.45 &7.71 &5.13 &4.43 &3.51 &5.19 \\
				
				CAG~\cite{zhang2019category}   & 53.98 &71.03 &77.38 &51.35 &63.44 &6.86 &3.27 &4.12 &3.87 &4.53 \\
				
				IntraDA~\cite{pan2020unsupervised}  & 61.38 &60.26 &70.48 &46.32 &59.61 &12.33 &9.26 &6.13 &5.72 &8.36 \\
				
				SIFAv2~\cite{chen2020unsupervised} & \textbf{65.30} &62.30 &78.90 &47.30 &63.40 &7.30 &7.40 &3.80 &4.40 &5.70\\
				
				MPSCL   & 64.66 & \textbf{77.34} & \textbf{81.61} &\textbf{55.90} &\textbf{69.87} &\textbf{5.59} & \textbf{2.64} &\textbf{3.44} & \textbf{3.50} &\textbf{3.80}\\
				
				\bottomrule[1.0pt]
		\end{tabular}}
	\end{center}
	\label{table:cardiac_seg}
\end{table}
\begin{figure}[!tbp]
	
	\begin{center}
		\includegraphics[width=0.42\textwidth ]{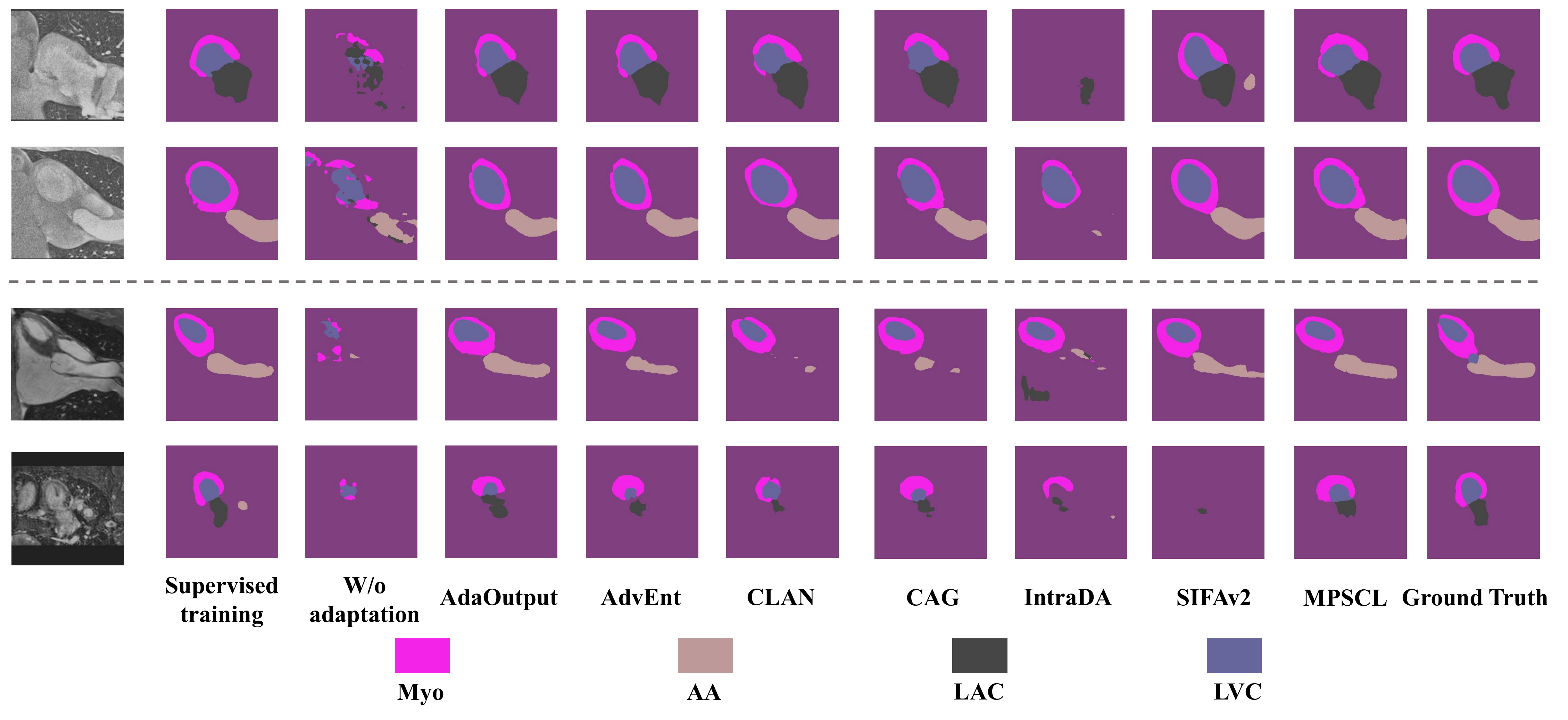}
	\end{center} 
	\caption{Visual comparison of segmentation results produced by different methods for cardiac CT slice images (1st-2nd row) and MRI slice images (3rd-4th row). From left to right are the raw test slice images (1st column), "Supervised training" upper bound (2nd column), "W/o adaptation" lower bound (3rd column), results of other unsupervised domain adaptation methods (4th-9th column), results of our MPSCL network (10th column), and ground truth (last column). The color corresponding to the semantic category is at the bottom.}
	\label{fig:slice_segmentation}
	
\end{figure}
\subsection{Compared Methods}
To evaluate the superiority of the proposed MPSCL, we compare with a wide range of UDA methods in cross-modal medical segmentation tasks.
These methods can be divided into two categories:
(i) \textit{\textbf{Category-agnostic global alignment}} {methods}. In this paradigm, we select two image-level alignment UDA methods (SIFAv1~\cite{chen2019synergistic} and SIFAv2~\cite{chen2020unsupervised}) and two methods (AdaOutput~\cite{tsai2018learning} and AdvEnt~\cite{vu2019advent}) which bridge the gap at the output-level;
(ii) \textit{\textbf{Category-aware local alignment}} {methods}. In view of this aspect, we focus on three category-aware alignment methods, including
category-level adversarial network(CLAN~\cite{luo2019taking}) based on self-adaptive adversarial loss, and two self-training based alignment methods (CAG~\cite{zhang2019category} and IntraDA~\cite{pan2020unsupervised}).
For a fair comparison, the generator architectures used for the implementation of other methods are the same as MPSCL except SIFAv1 and SIFAv2.

Since the same dataset and data preprocessing are applied for SIFAv1 and SIFAv2, as well as our model, the results from their papers are directly reported.
Additionally, the architectures of discriminator follows the configuration of PatchGAN~\cite{isola2017image}.

\subsection{Effectiveness of our MPSCL Model on Unsupervised Domain Adaptation}
In order to evaluate the importance of domain adaptation in cross-domain semantic segmentation, we first get the lower bound performance `without adaptation' ( named as W/o adaptation) by training a model only on source domain and directly generalizing it to the target domain.
In addition, we also provide the upper bound performance by conducting supervised learning on target domain to evaluate how much the gap is decreased between the `without adaptation' model and fully-supervised model.
For a fair comparison, the generator $G$ in our MPSCL framework is utilized for training the lower and upper bound models.

Table~\ref{table:cardiac_seg} presents the results for cross-modal cardiac segmentation task, including lower and upper models and several state-of-the-art UDA methods.
It can be seen that: i) the W/o adaptation model trained on MRI images only obtains the average Dice of 23.28$\%$ when being applied to CT domain.
Similarly, the model trained on CT images also achieves merely the average Dice of 20.35$\%$ on MRI domain.
These results are far below the performances 90.40$\%$ and 85.07$\%$ of supervised training models, which demonstrates the serious domain shift problem between MRI and CT domains;
ii) remarkably, our MPSCL model has achieved significant performance improvements in terms of both Dice and ASD measurements.
For CT images,  we improve the average Dice to 84.08$\%$ over the four cardiac structures with the average ASD being reduced to 3.47, and for MRI images, we obtain the average Dice of 69.87$\%$ and the average ASD 3.80;
iii) meanwhile, our method significantly outperforms other category-agnostic alignment methods by a large gain.
This shows that it is of great importance to maintain the semantic consistency between the two domains.
In addition, compared with category-aware alignment methods, our method also achieves significant improvements. 
For example, for CT images, our MPSCL achieves a clear improvement of 4.68$\%$ in the average Dice and a obvious reduction of 1.75 in the average ASD.
This shows that it is of great importance to enhance inter-class separability and intra-class compactness between the two domains.

\begin{figure*}[htbp]
	
	\begin{center}
		\includegraphics[width=0.7\textwidth]{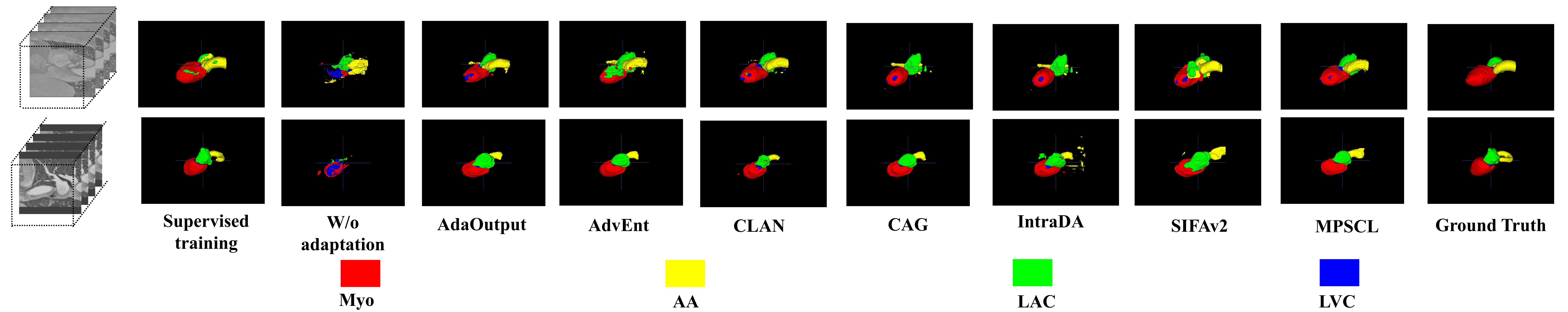}
	\end{center} 
	
	\caption{Visual comparison of segmentation results produced by different methods for cardiac CT data volume (1st rows) and MRI data volume (2nd rows). From left to right are the raw test volumes (1st column), "Supervised training" upper bound (2nd column), "W/o adaptation" lower bound (3rd column), results of other unsupervised domain adaptation methods (4th-9th column), results of our MPSCL network (10th column), and ground truth (last column). The color corresponding to the semantic category is at the bottom. }
	\label{fig:volume_segmentation}
	
\end{figure*}

Fig.~\ref{fig:slice_segmentation} presents some segmentation results of several examples, and it is obvious that the W/o adaptation model is hard to completely capture the correct cardiac structures due to the domain shift problem between the two domains.
Meanwhile, with comparison to the other UDA methods, the outputs of our MPSCL are more consistent with the ground truth for the slice images in both two transferring directions.
In addition, considering the practical clinical application, we also provide the 3D segmentation results of a patient volume data in Fig.~\ref{fig:volume_segmentation}.

Although our MPSCL is trained in a 2D view without considering correlation of inter-frames, very complete and accurate heart structure segmentation for CT and MRI volumes can also be obtained.
Both the qualitative and visualization results demonstrate that our MPSCL can effectively tackle the domain shift problem.

\begin{figure}[!tbp]
	
	\begin{center}
		\includegraphics[width=0.25\textwidth ]{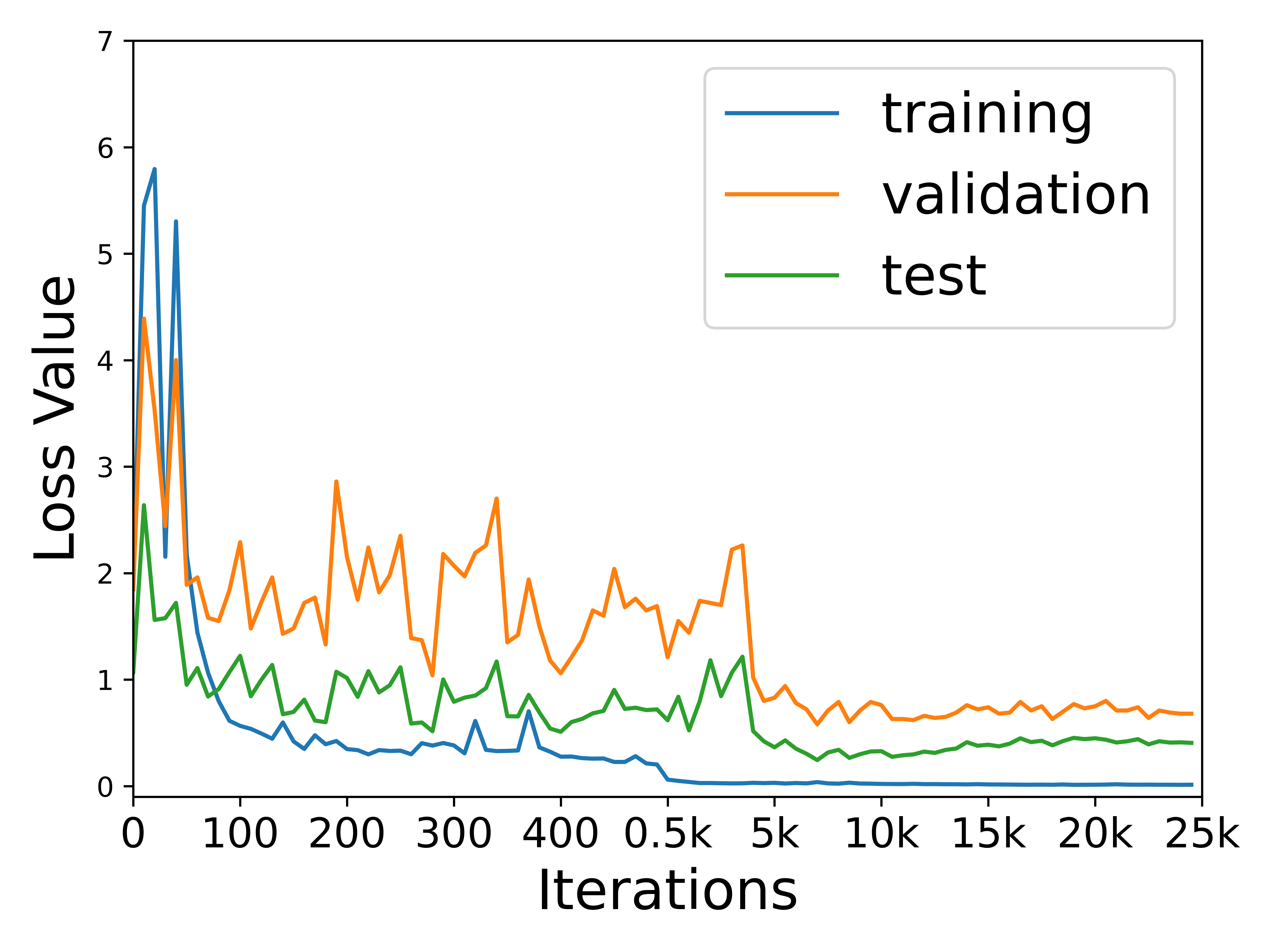}
	\end{center} 
	
	\caption{The loss curves of model training in MRI to CT alignment}
	\label{fig:loss}
	
\end{figure}

In addition, we also provide a Fig.~\ref{fig:loss} with the training/validation/test loss of weighted cross-entropy to examine it for overfitting effects. 
The above loss are calculated on the source training/target training/target test datasets, respectively.
It can be seen that, in the early stage of model training, the jitter of the loss function is more pronounced in the target validation and test datasets, but for the training set, it tends to converge quickly. 
This is because the discriminator can easily distinguish the source domain image and the target domain image at the beginning of training, resulting in the generator not learning a good domain-invariant representation. 
In addition, it can be seen that the loss variation trends in the validation and test datasets are the same during the training process, which means that our model is not overfitted during the training process. 
In particular, we also apply an early-stopping strategy to obtain the best model.
\subsection{Angle Distribution of Target Domain}
\begin{figure}[tbp]
	\begin{center}
		\includegraphics[width=0.37\textwidth ]{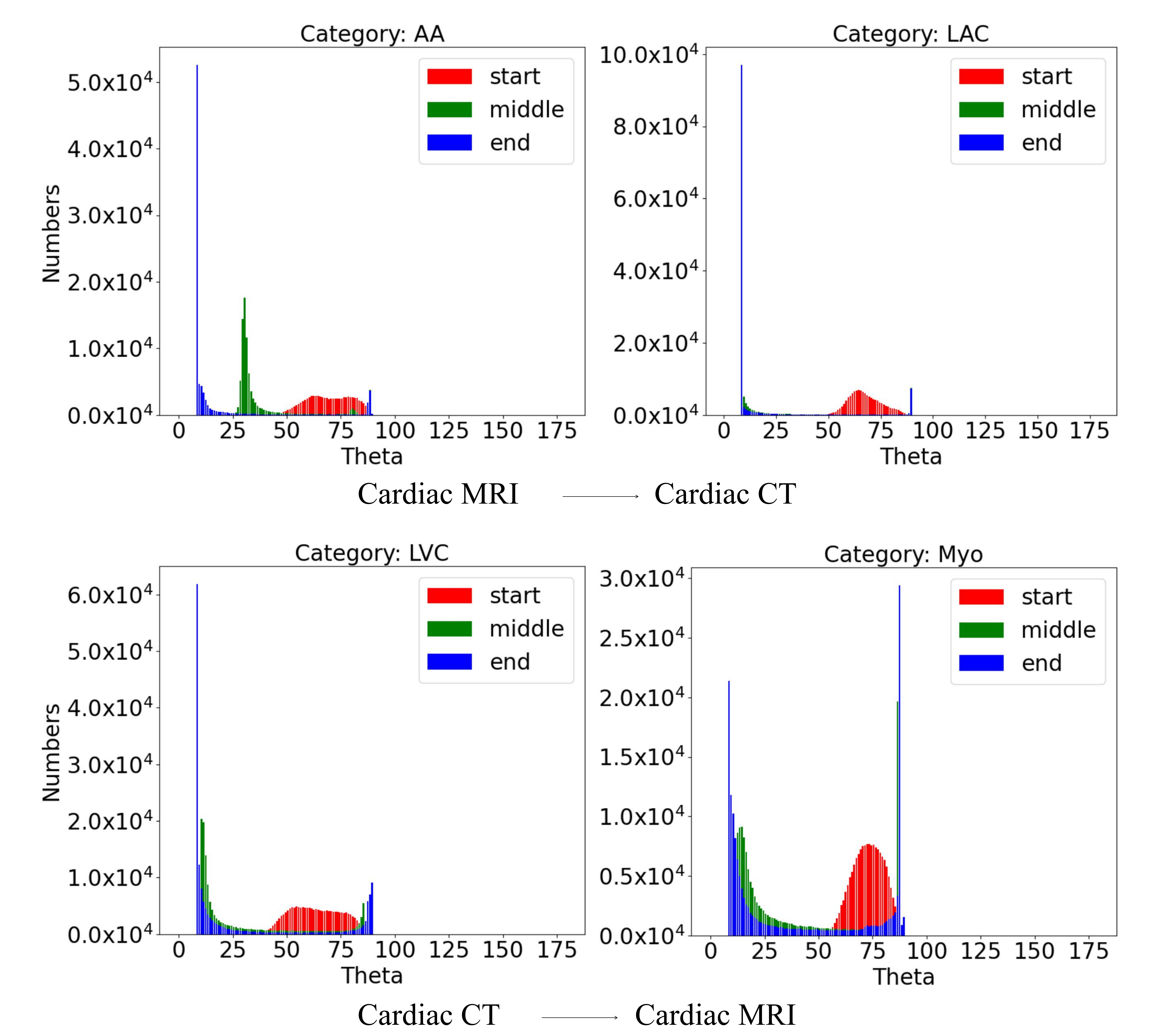}
	\end{center} 
	\caption{The distributions of the angle $\theta^{(k)}_{n}[\cdot;y_l]$ of target domain from the begining to end during MPSCL training. The top two are based on the application of MRI $\rightarrow$ CT , and the bottom two are from the setting of CT $\rightarrow$ MRI.}
	\label{fig:theta_distri}
\end{figure}

In the following, we demonstrate the effectiveness of the deviation angle penally in margin preserving contrastive loss given in Eq.(\ref{loss:MPCL}) from the view of statistic analysis.
Fig.~\ref{fig:theta_distri} presents the distributions of the angle $\theta^{(k)}_{n}[\cdot;y_{l}]$ of target domain.
It can be observed that, from the beginning to end of MPSCL training, the similarity between the pixel feature and the positive category anchor improve continuously, which means the gap between the two domains is gradually reduced.
In other words, the semantic consistency between the two domains is well preserved in our MPSCL.
What deserves noting is, at the end of the model training, most of the angles are concentrated in a smaller intervals.
Obviously, the intra-category compactness is enhanced. Meanwhile, due to the part of generated noisy pseudo-labels, some pixel feature have lower similarities with the positive category anchor.

\begin{table}[!tbp]
	\tiny
	\caption{Effectiveness of category-aware feature alignment.}
	\centering
	\begin{center}
		\resizebox{0.4\textwidth}{!}{%
			\begin{tabular}{c|ccccc|ccccc}
				\toprule[1.0pt]
				
				\multicolumn{11}{c}{Cardiac MRI $\rightarrow$ Cardiac CT}\\	
				\hline
				\multirow{2}{*}{Method} &\multicolumn{5}{c|}{Dice $\uparrow$}&\multicolumn{5}{c}{ASD$\downarrow$}\\
				\cline{2-11}
				&AA &LAC &LVC &MYO &Average &AA &LAC &LVC &MYO &Average \\
				
				\hline
				
				Baseline  &89.01 & \textbf{88.10} & 84.15 & 65.19 & 81.75 & 4.79 & \textbf{2.98} & 3.13 & 3.80 & 3.68\\
				
				MPSCL  &\textbf{90.26} & 87.08 & \textbf{86.45} &\textbf{72.51} &\textbf{84.08} &\textbf{3.47} &3.16 &\textbf{2.85} &\textbf{3.41} &\textbf{3.47}\\
				
				\bottomrule[1.0pt]
		\end{tabular}}
	\end{center}
	
	\centering
	\begin{center}
		\resizebox{0.4\textwidth}{!}{%
			\begin{tabular}{c|ccccc|ccccc}
				\toprule[1.0pt]
				
				\multicolumn{11}{c}{Cardiac CT $\rightarrow$ Cardiac MRI}\\	
				\hline
				\multirow{2}{*}{Method} &\multicolumn{5}{c|}{Dice $\uparrow$}&\multicolumn{5}{c}{ASD$\downarrow$}\\
				\cline{2-11}
				&AA &LAC &LVC &MYO &Average &AA &LAC &LVC &MYO &Average \\
				
				\hline
				
				Baseline  &\textbf{65.05} & 70.60 & 77.88 & 52.31 & 66.46 & \textbf{5.53} & 3.67 & 4.07 & 4.01 & 4.32\\
				
				MPSCL   & 64.66 & \textbf{77.34} & \textbf{81.61} &\textbf{55.90} &\textbf{69.87} &5.59 & \textbf{2.64} &\textbf{3.44} & \textbf{3.50} &\textbf{3.80}\\
				
				\bottomrule[1.0pt]
		\end{tabular}}
	\end{center}
	\label{table:ablation}
\end{table}

\subsection{Self-paced Pseudo-labels Visualization}
To provide more in-depth and visual examination of the self-paced pseudo-labels, we conduct a qualitative analysis as illustrated in Fig.~\ref{fig:pl_process}\footnote{In the supplementary material, we also provide the dynamic evolution of the self-paced pseudo-label assignment with model training going on.}.
Clearly, the generated pseudo-labels can provide informative supervision for conducting contrastive learning in target domain without human annotations. Our MPSCL model can progressively refine the pseudo-labels to correct the error and produce better supervision during model training in an `easy-to-hard' scheme.
At the start of model training, since the confidence difference $R^{(1)}_n$ between the maximum and submaximum confidence scores for each pixel region is not noticeable, only a few well-adapted pixel regions are selected although the noisy predictions are also removed.
But as training proceeds, the generated predictions come gradually closer to the ground-truth labels.
On the other hand, for the pixel regions poorly adapted before, they will receive more significant confidence difference.
With such self-pacing procedure, the generated pseudo-labels can provide more information to help generate better results on target categories.
However, it also should be noticed that the self-paced pseudo-labels also contain  incorrect information, which will lead to the negative transfer.

\begin{figure}[!htbp]
	\begin{center}
		\includegraphics[width=0.35\textwidth ]{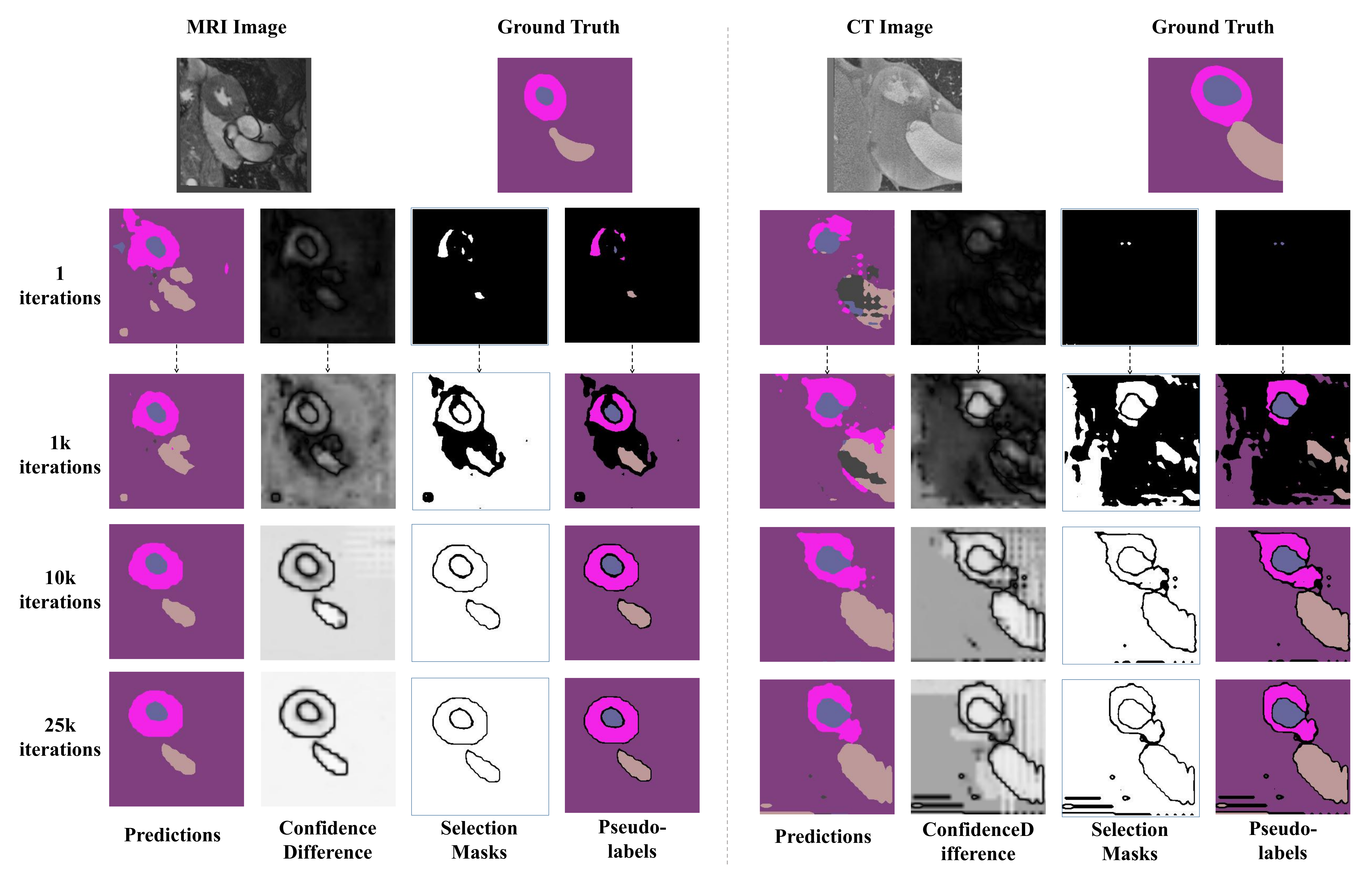}
	\end{center} 
	\caption{Visualization of pseudo-labels that are gradually refined during model training. The left side of the dotted line is a MRI slice and the right side is a CT slice. For each side, the first column is the generated predictions based on the domain-adaptive prototypes. The second column is the confidence difference between the maximum and the submaximum confidence scores. The third column is the pixel-level masks for interest region selection, where white indicates that the pixels are selected and black indicates unselected pixels. The fourth column is the generated pseudo-labels.}
	\label{fig:pl_process}
\end{figure}

\subsection{Ablation Analysis}
We conduct ablation experiments to evaluate the effectiveness of preserving the semantic consistency between the two domains in our method.
The ablation study results are show in Table.~\ref{table:ablation} containing MRI to CT and CT to MRI two applications, and our Baseline method in Table. ~\ref{table:ablation} only achieves global marginal feature alignment by setting $\beta = 0.0$.
It is obvious that our MPSCL improves the segmentation performance to a large degree, and for CT images, the average Dice is increased from 81.75$\%$ to 84.08$\%$, and the average ASD is reduced from 3.68 to 3.47.
Similarly, for MRI images, the average Dice is increased to 69.87$\%$ and the average ASD is reduced to 3.80.
This is due to the fact that our MPSCL significantly promotes the category-level alignment between the two domains and avoids the semantic confusion in feature space.
In addition, it is also worth noticing that there also exits slight degradation for some categories (i.e., LAC category in CT domain and AA category in MRI domain).
This maybe because that the threshold $\delta_{th}$ is not suitable for these categories, which results in the generated pseudo-labels containing more error information.

\subsection{Parameter Analysis}
\begin{figure}[!tbp]
	\begin{center}
		\includegraphics[width=0.3\textwidth]{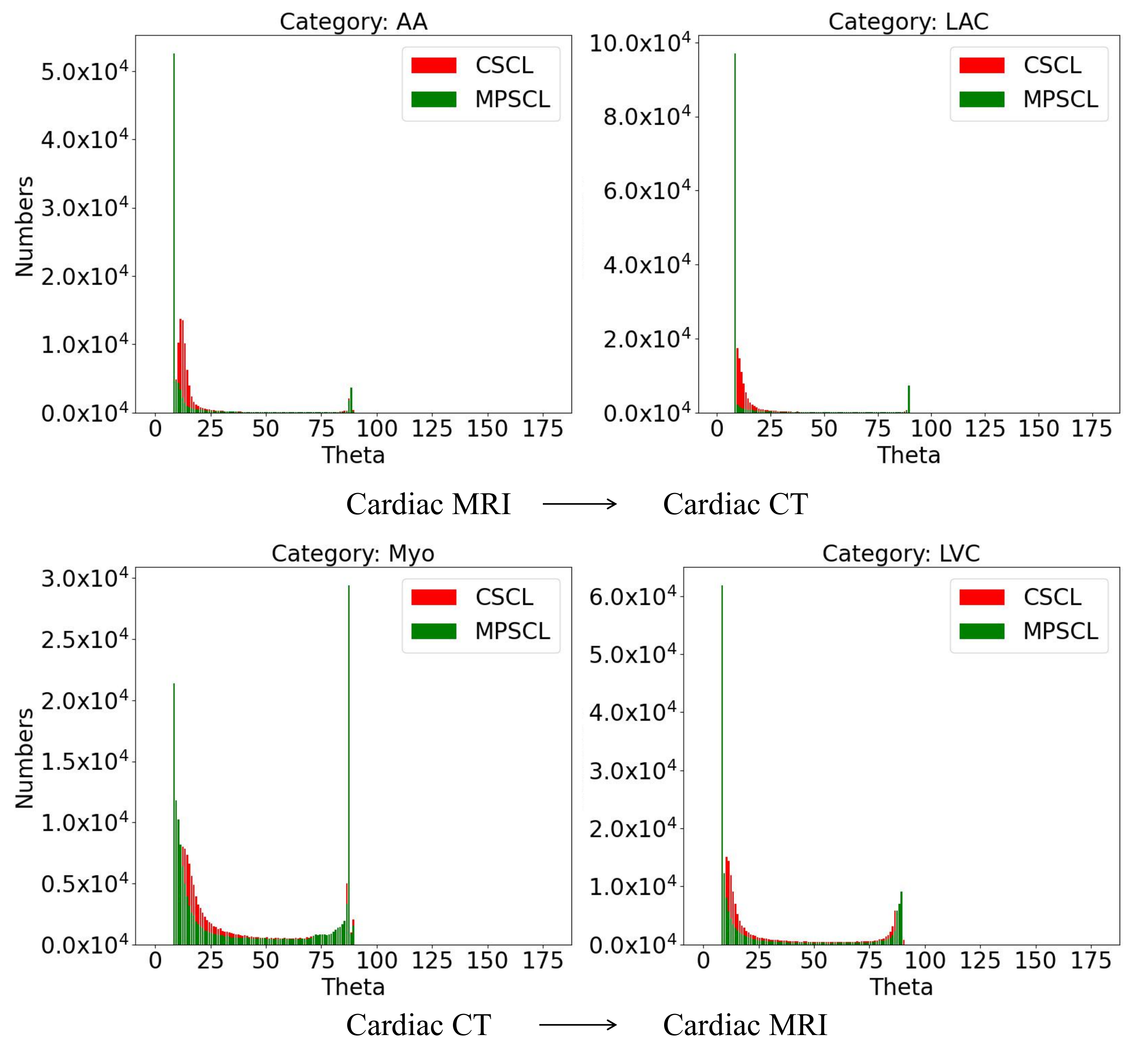}
	\end{center} 
	\caption{The distributions of the angle $\theta^{(k)}_{n}[\cdot;y_l]$ of target domain w.r.t. MPSCL and CSCL methods. }
	\label{fig:theta_diff_model}
\end{figure}

We also validate the influence of angular margin penalty in our MPSCL and the quantitative results are included in Table.~\ref{table:margin}, in which CSCL denotes the conventional self-paced contrastive learning without adopting the deviation angle penalty (i.e., $m=0.0$).
Meanwhile, we also present the angle distributions of different settings in Fig.~\ref{fig:theta_diff_model}. It can be observed that the average Dice is improved from 83.85$\%$ to 84.08$\%$, and the average ASD is decreased from 3.68 to 3.47 in the MRI to CT direction.
Additionally, for MRI images, our method improves the average Dice by more than 1.41$\%$ and reduces the average ASD by more than 0.27.

As illustrated in Fig.~\ref{fig:theta_diff_model}, the intra-category compactness is further enhanced by explicitly putting the deviation angle penalty between each pixel feature and the positive category anchor.
Both the qualitative and visualization results demonstrate that our MPSCL can further boost the discriminability of representation compared to CSCL.
It also should be pointed out, our model does not perform as well as CSCL in some categories (including LAC and LVC categories in CT domain and AA category in MRI domain).
The possible reason may be the pseudo-labels are not correctly assigned in theses cases, thus the deviation angle penalty may further enlarge the difference between the representations and the positive category anchor.

\begin{table}[!tbp]

	\caption{The influence of deviation angle penalty $m$.
	}
	\centering
	\begin{center}
		
		\resizebox{0.4\textwidth}{!}{%
			\begin{tabular}{c|ccccc|ccccc}
				\toprule[1.0pt]
				
				\multicolumn{11}{c}{Cardiac MRI $\rightarrow$ Cardiac CT}\\	
				\hline
				\multirow{2}{*}{Method} &\multicolumn{5}{c|}{Dice $\uparrow$}&\multicolumn{5}{c}{ASD$\downarrow$}\\
				\cline{2-11}
				&AA &LAC &LVC &MYO &Average &AA &LAC &LVC &MYO &Average \\
				
				\hline
				
				CSCL  &89.91 & \textbf{87.96} & \textbf{86.80} & 70.73 & 83.85 & 5.38 & \textbf{2.87} & \textbf{2.84} & 3.44 & 3.63\\
				
				MPSCL  &\textbf{90.26} & 87.08 & 86.45 &\textbf{72.51} &\textbf{84.08} &\textbf{3.47} & 3.16 & 2.85 &\textbf{3.41} &\textbf{3.47}\\
				
				\bottomrule[1.0pt]
		\end{tabular}}
	\end{center}
	
	\centering
	\begin{center}
		\resizebox{0.4\textwidth}{!}{%
			\begin{tabular}{c|ccccc|ccccc}
				\toprule[1.0pt]
				
				\multicolumn{11}{c}{Cardiac CT $\rightarrow$ Cardiac MRI}\\	
				\hline
				\multirow{2}{*}{Method} &\multicolumn{5}{c|}{Dice $\uparrow$}&\multicolumn{5}{c}{ASD$\downarrow$}\\
				\cline{2-11}
				&AA &LAC &LVC &MYO &Average &AA &LAC &LVC &MYO &Average \\
				
				\hline
				
				CSCL  &\textbf{67.06} & 74.07 & 78.62 & 54.07 & 68.45 & \textbf{5.44} & 2.90 & 3.83 & 4.11 & 4.07\\
				
				MPSCL   & 64.66 & \textbf{77.34} & \textbf{81.61} &\textbf{55.90} &\textbf{69.87} &5.59 & \textbf{2.64} &\textbf{3.44} & \textbf{3.50} &\textbf{3.80}\\
				
				\bottomrule[1.0pt]
		\end{tabular}}
	\end{center}
	\label{table:margin}
\end{table}

\section{Conclusions}
For the challenge of unsupervised domain adaptation in medical image segmentation, an innovative MPSCL model is proposed, which promotes category-aware feature alignment by cross-domain contrastive learning.
To the best of our knowledge, this is the first time to introduce contrastive learning for this practical problem.
Specially, the domain-adaptive category prototypes are exploited to constitute contrastive pairs for the joint contrastive learning between the two domains.
We generate informative self-paced pseudo-labels for target domain to perform contrastive learning in target domain without prior label available.
The discriminability of representation is boosted by a margin preserving contrastive learning loss.
It is worth noticing that one category prototype per category does not cover the overall distribution.
Thus, our ongoing research work includes learning prototypes adaptively with the data distribution.

\section*{Acknowledgment}

This work was supported in part by Science and Technology Innovation 2030 - "New Generation Artificial Intelligence" Major Project under Grant No.2018AAA0102101, in part by the National Natural Science Foundation of China under Grant No.61976018.

\bibliographystyle{IEEEtran}      
\bibliography{mybib}                        

\end{document}